\documentclass[letterpaper, 10 pt, conference]{ieeeconf}  

\IEEEoverridecommandlockouts                              

\usepackage{amsmath} 
\usepackage{amsthm}  
\usepackage{amsfonts}
\usepackage{amssymb}  
\usepackage{dsfont}
\usepackage[dvipsnames]{xcolor}
\usepackage{tikz}
\usepackage{pgfplots}
\usepackage[hidelinks]{hyperref}
\pgfplotsset{compat=1.18,
	/pgfplots/ybar legend/.style={
		/pgfplots/legend image code/.code={%
			\draw[##1,/tikz/.cd,yshift=-0.25em]
			(0cm,0cm) rectangle (3pt,0.8em);},
}}
\usepackage[nolist,nohyperlinks]{acronym}
\usepackage[capitalize]{cleveref}
\Crefname{figure}{Fig.}{Figs.}
\crefname{equation}{}{}
\Crefname{equation}{Equation}{Equations}

\usepackage[per-mode=symbol,exponent-product=\cdot]{siunitx}

\usepackage{url}
\usepackage[numbers]{natbib}
\usepackage{letltxmacro}
\LetLtxMacro{\autocite}{\citep}
\LetLtxMacro{\textcite}{\citet}

\usepackage{booktabs}
\usepackage{tabularx}
\newcolumntype{Y}{>{\raggedright\arraybackslash}X}
\newcommand{\otoprule}{\midrule[\heavyrulewidth]}

\let\originalleft\left
\let\originalright\right
\renewcommand{\left}{\mathopen{}\mathclose\bgroup\originalleft}
\renewcommand{\right}{\aftergroup\egroup\originalright}
\newcommand{\actor}{a}
  \newcommand{\actorset}{\mathcal{A}}
  \newcommand{\actorscenariosset}{\mathcal{B}}
  \newcommand{\numberofscenariosactortime}[2]{K\left(#1,#2\right)}
\newcommand{\actorcoverage}[1]{\coveragesymbol_{\mathrm{A}}\left(#1\right)}
\newcommand{\cardinality}[1]{\left|#1\right|}
\newcommand{\categorysymbol}{C}
  \newcommand{\category}[1]{\categorysymbol_{#1}}
  \newcommand{\categoryset}{\mathcal{C}}
\newcommand{\coveragesymbol}{\mathrm{Coverage}}
\newcommand{\numbervar}{n}
\newcommand{\setpositiveint}{\mathds{Z}^+}
\newcommand{\tagsymbol}{L}
  \newcommand{\tagvar}[1]{\tagsymbol_{#1}}
  \newcommand{\tagset}{\mathcal{L}}
  
  \newcommand{\tagcount}[2]{N\left(#1,#2\right)}
\newcommand{\tagcoverage}[1]{\coveragesymbol_{\text{Tag}}\left(#1\right)}
\newcommand{\timevar}{t}
  \newcommand{\timeset}{\mathcal{T}}
  \newcommand{\timeactorset}[1]{\timeset_{#1}}
  \newcommand{\numberofscenarios}[1]{M\left(#1\right)}
\newcommand{\timecoverage}[1]{\coveragesymbol_{\mathrm{T}}\left(#1\right)}
\newcommand{\timeactorcoverage}[1]{\coveragesymbol_{\mathrm{AT}}\left(#1\right)}

\makeatletter
\newcommand*{\org@overidelabel}{}
\let\org@overridelabel\@verridelabel
\renewcommand*{\@verridelabel}[1]{%
	\@bsphack
	\protected@write\@auxout{}{\string\AC@undonewlabel{#1@cref}}%
	\org@overridelabel{#1}%
	\@esphack
}%
\makeatother

\begin{acronym}[AAAAAAAA]
	\acro{alks}[ALKS]{Automated Lane Keeping System}\acroindefinite{alks}{an}{an}
	\acro{ads}[ADS]{Automated Driving System}\acroindefinite{ads}{an}{an}
	\acro{ccam}[CCAM]{Connected, Cooperative, and Automated Mobility}
    \acro{natm}[NATM]{New Assessment/Test Methods}
	\acro{odd}[ODD]{Operational Design Domain}\acroindefinite{odd}{an}{an}
	\acro{saf}[SAF]{Safety Assessment Framework}
	\acro{sc}[SC]{Scenario Category}\acroindefinite{sc}{an}{a}\acrodefplural{sc}[SCs]{Scenario Categories}
	\acro{tod}[TOD]{Target Operational Domain}
	\acro{unece}[UNECE]{United Nations Economic Commission for Europe}
\end{acronym}

\newlength{\figurewidth}
\newlength{\figureheight}

\title{\LARGE \bf
Coverage Metrics for a Scenario Database for the Scenario-Based Assessment of Automated Driving Systems
}

\author{Erwin de Gelder$^{1*}$, Maren Buermann$^{1}$, Olaf Op den Camp$^{1}$
\thanks{The research presented in this work has been made possible by the SUNRISE project.
	This project is funded by the European Union's Horizon Europe Research \& Innovation Actions under grant agreement No.\ 101069573. 
	Views and opinions expressed are however those of the authors only and do not necessarily reflect those of the European Union or the European Climate, Infrastructure and Environment Executive Agency (CINEA). 
	Neither the European Union nor the granting authority can be held responsible for them.}
\thanks{$^{1}$TNO, Integrated Vehicle Safety, Helmond, The Netherlands}%
\thanks{$^{*}$Corresponding author: {\tt\small erwin.degelder@tno.nl}}%
}

\newcommand{\cstart}{}
\newcommand{\cend}{}

\begin{document}

	\maketitle
	\thispagestyle{empty}
	\pagestyle{empty}

	\begin{abstract}
	
 
    \Acp{ads} have the potential to make mobility services available and safe for all. 
    A multi-pillar \ac{saf} has been proposed for the type-approval process of \acp{ads}. 
    The \ac{saf} requires that the test scenarios for the ADS adequately covers the \ac{odd} of the \ac{ads}. 
    A common method for generating test scenarios involves basing them on scenarios identified and characterized from driving data.
     
    This work addresses two questions when collecting scenarios from driving data. 
    First, do the collected scenarios cover all relevant aspects of the \ac{ads}' \ac{odd}? 
    Second, do the collected scenarios cover all relevant aspects that are in the driving data, such that no potentially important situations are missed? 
    This work proposes coverage metrics that provide a quantitative answer to these questions. 

    The proposed coverage metrics are illustrated by means of an experiment in which over \num{200000} scenarios from \num{10} different scenario categories are collected from the HighD data set. 
    The experiment demonstrates that a coverage of \SI{100}{\%} can be achieved under certain conditions, and it also identifies which data and scenarios could be added to enhance the coverage outcomes in case a \SI{100}{\%} coverage has not been achieved. 
    Whereas this work presents metrics for the quantification of the coverage of driving data and the identified scenarios, this paper concludes with future research directions, including the quantification of the completeness of driving data and the identified scenarios.
    
\end{abstract}
    \acresetall
	\section{INTRODUCTION}
\label{sec:introduction}

The road traffic system is changing rapidly, due to changes in the existing mobility system (e.g., the increasing share of cycling), the introduction of new mobility systems such as \ac{ccam} systems, shared mobility concepts, and new enabling technologies such as artificial intelligence and wireless V2X-communication \autocite{piao2008advanced}. 
Simultaneously, vehicles are increasingly automated. 
The goal of automation is to make mobility services available and safe for all, including vulnerable road users. 
At the same time, automation can provide more comfort to drivers and passengers and increase the efficiency of the mobility system.

Authorities are being asked to allow vehicles equipped with new advanced communication and automation technologies onto public roads. 
To put a legal framework to the safe deployment of \acp{ads}, regulations are being implemented by the \ac{unece}, e.g., for \acp{alks} \autocite{ece2022WP29}, and the European Commission, i.e., for \ac{ccam} systems in four use cases \autocite{ec2022typeapprovalads}. 
The \ac{unece} WP.29 Working Party on Automated/Autonomous and Connected Vehicles (GRVA) has developed the \ac{natm} Master Document \autocite{ece2021natm}, in which a multi-pillar \ac{saf} is proposed for the type-approval process of \ac{ccam} systems.

Though \acp{ads} might be complex, and the assessment procedure of such systems might be complicated, the assessment results should be unambiguous, easily understood by experts in the field, and explainable to authorities and the public. 
This is one reason that scenarios, as a structured way to describe the large varieties of situations and conditions that \iac{ads} may encounter on the road, form the most important source of information to generate test scenarios for the different ways of testing: virtual testing using computer models and simulation tools, track testing under realistic and reproducible conditions, and real-world tests on the road, e.g., by means of field operational tests.

A common way of collecting (naturalistic driving) data for the identification and characterization of scenarios is by the use of instrumented vehicles driving on public roads \autocite{elrofai2018scenario}.
In such data collection efforts, the ego vehicle refers to the vehicle that is perceiving the world through its sensors or the vehicle that must perform a specific task. 
A scenario describes any situation on the road encountered by the ego vehicle and how this situation develops over time. 
A drive on the road is considered a continuous sequence of scenarios --- which might overlap.

An important characteristic of \iac{ads} is represented by its \ac{odd}, which refers to the ``operating conditions under which a given driving automation system or feature thereof is specifically designed to function, including, but not limited to, environmental, geographical, and time-of-day restrictions, and/or the requisite presence or absence of certain traffic or roadway characteristics'' \autocite{sae2021j3016}. 
In other words, the \ac{odd} is used to describe the external environment and conditions for which \iac{ads} system is designed for. 
The importance of describing the \ac{odd} is underlined by the different initiatives for specifying \iac{odd}, e.g., see \autocite{ISO34503, BSI1883-2020, openodd}. 
Since scenarios are a description of the environment of the \ac{ads}-equipped vehicle, scenarios can be used to describe the \ac{odd}.

The trustworthiness of the safety assessment results of \iac{ads} depends on the quality of the selection of test scenarios, and consequently on determining the range of scenarios that is relevant for the \ac{ads} under assessment. 
Hence, trustworthiness depends on how well the underlying data for scenario identification and collection as well as the selected set of scenarios cover the \ac{odd}. 
In this paper, we study metrics to quantify how well observed scenarios cover \iac{odd}.

Coverage can be measured in multiple ways, which is why this work presents multiple coverage metrics. 
The first coverage metric quantifies whether all relevant aspects of \iac{odd} --- labeled with the use of tags --- are captured.
The other coverage metrics measure the coverage of the data containing scenarios with respect to time, actors, and a combination of time and actors.
To illustrate the use of the proposed metrics, they have been implemented to determine the coverage for the StreetWise \acp{sc} on the HighD data set \autocite{krajewski2018highD}.
We demonstrate how the resulting numbers can be used to determine what data and which \acp{sc} may need to be included for increasing the trustworthiness of the safety assessment results.

After a short overview of related works in \cref{sec:related works}, the different coverage metrics are introduced in \cref{sec:metrics}. 
Results of the experiment are provided in \cref{sec:results}. 
After discussion of the results in \cref{sec:discussion}, conclusions are drawn and recommendations for further research are given in \cref{sec:conclusions}.

	\section{RELATED WORKS}
\label{sec:related works}

The term ``coverage'' is informally understood as the degree to which something deals with something else. 
For example, in statistics, the coverage of a confidence interval of $p$ indicates the actual probability that the confidence interval contains $p$ \autocite{wasserman2004statistics}. 
In the field of software engineering, coverage is ``a measure of verification and completeness'' \autocite{piziali2007functional} and, as described by \textcite{piziali2007functional}, there is no single (best) way to define coverage.
Coverage metrics can be tailored to the verification progress from the perspective of the functional requirements of the product (functional coverage), the part of the code that have been executed during the verification (code coverage), the fractions of assertions evaluated (assertion coverage), etc.
To illustrate the many ``coverage'' metrics, consider the code coverage. 
The code coverage can be measured in terms of the lines of codes that have been executed, the branches or the paths, etc.

In \autocite{alexander2015situation}, the importance of coverage metrics is emphasized for the testing of autonomous vehicles, as the authors argue that testing that inadequately covers the situations that an autonomous vehicle will encounter is similar to inadequate testing.
Therefore, \textcite{alexander2015situation} have proposed the use of a ``situation coverage metric''.  
They have suggested that such a metric should be tractable, which has two implications.
First, a percentage should be calculated.
Thus, when considering \iac{ads}, the number of kilometers driven or the number of (simulated) scenarios are insufficient examples, because both can be infinite. 
Also, the number of failures found is not a good example, because the total number of failures is unknown.
Second, a \SI{100}{\percent} coverage should be achievable under realistic practical conditions. 

In the field of testing of \acp{ads}, the term ``coverage'' is often used as a measure used to decide the adequacy of a testing effort and as a stopping criteria for testing \autocite{araujo2023testing}.
\textcite{riedmaier2020survey} defined the term ``scenario coverage'' as the extent to which concrete scenarios used for testing cover the entire space, without further defining quantitative measures.
In \autocite{weissensteiner2023operational}, this idea is further exploited as several metrics are proposed for measuring the coverage of concrete scenarios with respect to the \ac{odd} of \iac{ads}.
Note that given the fact that the number of concrete scenarios is virtually infinite \autocite{amersbach2019defining}, and following the aforementioned reasoning of \textcite{alexander2015situation}, using concrete scenarios will not provide a good coverage metric.
As an alternative for concrete scenarios, the types of scenarios or the scenes may be considered, where a scene refers to the situation at a single time instant of a scenario.
Although \textcite{hauer2019didwe} did not mention the term ``coverage'', the metric that they proposed estimates the number of types of scenarios that are not addressed during testing.
In \autocite{alnaser2021autonomous}, a coverage metric is defined using scenes, although no practical results are presented.

When deriving test scenarios from scenarios observed in real-world data, the real-world data should provide good coverage. 
Compared with the amount of literature on coverage regarding the testing effort of \acp{ads}, there is little literature available regarding the coverage of the real-world data. 
In \autocite{wang2017much}, a criterion is proposed for the collection of naturalistic driving data. 
In \autocite{degelder2019completeness}, the asymptotic mean integrated squared error of an estimated probability density function is used as a metric to quantify the coverage of the collected data.
A disadvantage of both these works is that a \SI{100}{\percent} coverage cannot be achieved. 
In \autocite{hartjen2020saturation}, a metric is proposed based on the number of distinct sequences of maneuvers of an observed object. 
A disadvantage of this metric is that the total number of distinct sequences is unknown, so a percentage cannot be calculated.
Recently, \textcite{glasmacher2024towards} defined coverage with respect to a set of scenarios as ``the quantifiable extent to which a set of scenarios or parameters represent a defined \ac{odd} or predefined set of scenarios'', but no metric has been proposed as \autocite{glasmacher2024towards} focused on completeness instead (more on that in the discussion of \cref{sec:discussion}).
Glasmacher et al.\ proposed a coverage metric based on scenario parameter values in \autocite{glasmacher2023acquire}. 
However, this approach requires selecting a parameterization and limiting the number of parameters, as achieving \SI{100}{\percent} coverage could be impractical otherwise. 
Despite this drawback, their approach is promising and complements the method presented in this work.

	\section{COVERAGE METRICS}
\label{sec:metrics}

In this work and in line with \autocite{glasmacher2024towards}, coverage is defined as the degree to which a set of scenarios observed in real-world data cover \iac{odd}. 
To further distinguish the metrics that are proposed later in this section, two types of coverage are considered, both aiming  to answer different questions:
\begin{itemize}
	\item Type I: Do the collected scenarios cover all relevant aspects of \iac{odd}?
	\item Type II: Do the collected scenarios cover all relevant aspects that are in the driving data?
\end{itemize}
Four different coverage metrics are proposed.
The first metric is the tag-based coverage, which addresses coverage type I.
The other three metrics, i.e., time-based coverage, actor-based coverage, and actor-over-time-based coverage, address coverage type II.

quantifiable extent to which a set of scenarios or parameters represent an ODD

\subsection{Tag-based coverage}
\label{sec:tag coverage}

Before introducing the tag-based coverage metric, we need to distinguish scenarios from \acp{sc} \autocite{degelder2022ontology}.
Here, a scenario refers to a quantitative description of the relevant characteristics of the ego vehicle, its activities and/or goals, its static environment, and its dynamic environment. 
In contrast, \iac{sc} refers to a qualitative description of the ego vehicle, its activities and/or goals, its static environment, and its dynamic environment. 
For example, the \ac{sc} ``cut in'' comprises all possible cut-in scenarios. 
Scenarios may further be enriched with tags, e.g., a scenario belonging to the \ac{sc} ``cut in'' may have the tag ``actor at left'' to indicate that there is an actor at the left side of the ego vehicle that prevents the ego vehicle from changing lane to the left.

Let $\tagset$ denote a set of tags and let $\categoryset$ denote a set of \acp{sc}. 
Note that the set of tags should be based on the relevant aspects of \iac{odd}, whereas the set of \acp{sc} could be based on the coverage type II metrics presented in \cref{sec:time coverage,sec:actor coverage,sec:time actor coverage}.
For the tag-based coverage, we make use of the function $\tagcount{\tagsymbol}{\categorysymbol}$, which returns the number of scenarios that belong to \ac{sc} $\categorysymbol$ and contain the tag $\tagsymbol$.
Continuing the previous example, in case we have 10 cut-in scenarios with an actor at the left of the ego vehicle, we would have $\tagcount{\text{Actor at left}}{\text{Cut-in}}=10$.
The tag-based coverage metric is defined as follows:
\begin{equation}
	\label{eq:tag coverage}
	\tagcoverage{\numbervar}
	= \frac{1}{\numbervar \cardinality{\tagset} \cardinality{\categoryset}}
	\sum_{\tagsymbol \in \tagset} \sum_{\categorysymbol \in \categoryset}
	\min\left( \numbervar, \tagcount{\tagsymbol}{\categorysymbol} \right),
\end{equation}
where $\numbervar\in\setpositiveint$ and $\cardinality{\cdot}$ denotes the cardinality, e.g., $\cardinality{\tagset}$ equals the number of (distinct) tags.
In case $\tagcoverage{1}=1$, each tag is associated to at least one scenario of each \ac{sc}.

\cstart
For this coverage metric, three choices need to be made:
\begin{enumerate}
	\item The \acp{sc} belonging to $\categoryset$. 
	The \acp{sc} should cover the \ac{odd}. 
	The set of \acp{sc} could be based on relevant literature \autocite{USDoT2007precrashscenarios, degelder2019scenariocategories}, though we suggest using other coverage metrics to justify that the set of \acp{sc} is complete.
	As mentioned before, the metrics presented in \cref{sec:time coverage,sec:actor coverage,sec:time actor coverage} may be used.
	
	\item The tags belonging to $\tagset$.
	The tags should follow from the \ac{odd} description.
	When defining the \ac{odd} in accordance with the ISO~34503 standard \autocite{ISO34503}, the corresponding tags listed in the ISO~34504 standard \autocite{ISO34504} may be used.
	
	\item The required number of tags per \ac{sc}, $\numbervar$.
	Minimally, $\numbervar=\num{1}$, but to achieve more accurate statistics, if may be required to choose a higher value for $\numbervar$.
	To determine $\numbervar$, other metrics be used, e.g., see \autocite{wang2017much, degelder2019completeness, hartjen2020saturation, glasmacher2024towards, glasmacher2023acquire}.
\end{enumerate}
\cend

To obtain more accurate statistics of the scenarios belonging to \iac{sc}, it may be desired to have at least several scenarios of each \ac{sc} with a certain tag.
In that case, a larger value of $\numbervar$ may be chosen.

Note that different tag-based coverage metrics can be defined if different sets of tags are considered. 
For example, one may choose to calculate \cref{eq:tag coverage} with $\tagset$ consisting of tags related to environmental conditions, such as weather and lighting conditions, and with another set of tags consisting of scenery attributes, such as different types of roads.

\subsection{Time-based coverage}
\label{sec:time coverage}

The time-based coverage metric answers the question of whether all timestamp in the data is covered by one or more scenarios. 
Let $\timeset$ denote the set of all timestamps in the data set.
For the time-based coverage, we introduce the function $\numberofscenarios{\timevar}$, which returns the number of scenarios at time $\timevar$.
Note that it may be possible that scenarios happen in parallel, e.g., a leading vehicle decelerating and another vehicle overtaking the ego vehicle.
The time-based coverage metric is defined as follows:
\begin{equation}
	\label{eq:time coverage}
	\timecoverage{\numbervar}
	= \frac{1}{\numbervar\cardinality{\timeset}}
	\sum_{\timevar \in \timeset}
	\min\left( \numbervar, \numberofscenarios{\timevar} \right),
\end{equation}
with $\numbervar\in\setpositiveint$.
In case $\timecoverage{1}=1$, all timestamps in the data are covered by at least one scenario.
To account for the number of scenarios that can occur in parallel, one can increase the value of $\numbervar$.

\subsection{Actor-based coverage}
\label{sec:actor coverage}

The actor-based coverage metric answers the question of whether every relevant actor is covered by at least one scenario.
Let $\actorset$ denote the set of relevant actors. 
Here, the term ``relevant'' could be defined using some conditions.
For example, $\actorset$ could contain all actors that are at some point in time within a certain distance of the ego vehicle. 
Alternatively, $\actorset$ could contain all emergency vehicles in the data set, etc.
Let $\actorscenariosset$ denote the set of actors that are part of at least one scenario.
Then, the actor-based coverage metric is defined as follows:
\begin{equation}
	\label{eq:actor coverage}
	\actorcoverage{\actorset}
	= \frac{\cardinality{\actorset \cap \actorscenariosset}}{\cardinality{\actorset}}.
\end{equation}

\subsection{Actor-over-time-based coverage}
\label{sec:time actor coverage}

Achieving $\actorcoverage{\actorset}=1$ means that all actors of the set $\actorset$ are part of at least one scenario.
However, it does not consider the temporal aspect of when these actors are part of a scenario.
For example, it could be the case that an actor is near the ego vehicle --- and thus part of $\actorset$ --- but only part of a scenario once this vehicle is far away. 
To accommodate the time aspect, we introduce the fourth coverage metric; the actor-over-time-based coverage.

Let $\timeactorset{\actor}$ denote the set of timestamps at which the actor $\actor\in\actorset$ satisfies the conditions that makes this actor part of $\actorset$. 
Furthermore, let $\numberofscenariosactortime{\actor}{\timevar}$ be the number of scenarios at time $\timevar$ that contain actor $\actor$. 
Then, the actor-over-time-based coverage is defined as follows:
\begin{equation}
	\label{eq:time actor coverage}
	\timeactorcoverage{\actorset}
	= \frac{1}{\cardinality{\actorset}} \sum_{\actor\in\actorset}
	\frac{1}{\cardinality{\timeactorset{\actor}}} \sum_{\timevar\in\timeactorset{\actor}}
	\min\left( 1, \numberofscenariosactortime{\actor}{\timevar} \right).
\end{equation}

	\section{RESULTS}
\label{sec:results}

To illustrate the use of the coverage metrics that are presented in \cref{sec:metrics}, the metrics are evaluated based on scenarios from real-world data.
The setup of the experimental results are presented in \cref{sec:results setup} and the results are presented in the subsequent subsections.

\subsection{Setup experiment}
\label{sec:results setup}

The HighD data set \autocite{krajewski2018highD} is chosen for the experiment because of its size (more than \SI{40000}{\kilo\meter} of naturalistic driving data) and high accuracy.
The data consists of trajectories of cars and trucks at six different locations on German motorways obtained using video footage from drones.

To obtain the scenario data, each of the more than \num{100000} vehicles is treated as an ego vehicle once. 
I.e., from the total data set, more than \num{100000} smaller data sets are created, where each of the smaller data sets contains a single ego vehicle and trajectory data relative to the ego vehicle as if the other vehicles are perceived from the ego vehicle.
It is assumed that the ego vehicle can see all of its surrounding vehicles within a distance of \SI{100}{\meter}.
Each of the smaller data sets stops whenever the ego vehicle is \SI{100}{\meter} from its final position; this is done to avoid the sudden disappearance of vehicles in front of the ego vehicle, as these vehicles would be out of view of the drone camera.
Note that, as a result, vehicles with a trajectory less than \SI{100}{\meter} are not considered as ego vehicles.
In total, this resulted in \num{109986} data sets with a single ego vehicle.

\Cref{tab:scenario categories} lists the \num{10} scenario categories considered in this study.
This table also summarizes the activities of the ego vehicle and the main actor(s). 
Here, the main actor(s) refers to the actor(s) that are necessary for the scenario to occur.
That is, there may be other actors participating in the scenario as well, e.g., a vehicle overtaking the ego vehicle in the leading vehicle cruising scenario.
Based on the activities of the ego vehicle and the main actors and the approach outlined in \autocite{degelder2020scenariomining}, the scenarios are automatically extracted.
\Cref{tab:scenario categories} also indicates the number of scenarios found for each \ac{sc}.

\begin{table*}
	\centering
	\caption{Description of the \num{10} scenario categories that are considered in this work's experiment.}
	\label{tab:scenario categories}
	\begin{tabularx}{\textwidth}{l l l X l}
		\toprule
		Symbol & Name & Ego vehicle activity & Main actor(s) activity & Count \\ \otoprule
		$\category{1}$ & Leading vehicle cruising & Keeping lane & Keeping lane and cruising & 102308 \\
		$\category{2}$ & Leading vehicle accelerating & Keeping lane & Keeping lane and accelerating & 22296 \\
		$\category{3}$ & Leading vehicle decelerating & Keeping lane & Keeping lane and decelerating & 20351 \\
		$\category{4}$ & Approaching slower vehicle & Keeping lane & Keeping lane and driving slower than ego vehicle & 5052 \\
		$\category{5}$ & Cut-in in front of ego vehicle & Keeping lane & Changing lane to become leading vehicle & 2992 \\
		$\category{6}$ & Cut-out in front of ego vehicle & Keeping lane & Leading ego vehicle and then changing lane & 3069 \\
		$\category{7}$ & Changing lane with vehicle behind & Changing lane & Behind ego vehicle on adjacent lane & 2156 \\
		$\category{8}$ & Merging into an occupied lane & Changing lane & Both main actors stay in lane and become leading and following vehicles after ego vehicle lane change & 819 \\
		$\category{9}$ & Ego vehicle overtaking vehicle & Keeping lane & Keeping lane on overtaken by ego vehicle on adjacent lane & 38147 \\
		$\category{10}$ & Vehicle overtaking ego vehicle & Keeping lane & Keeping lane and overtaking ego vehicle on adjacent lane & 40307 \\
		\bottomrule
	\end{tabularx}
\end{table*}

\subsection{Results tag-based coverage}
\label{sec:results tag based}

For the tag-based coverage, \num{18} different tags are considered, see \cref{tab:tag matrix}.
The first two tags apply to the two different types of vehicles that are considered in the HighD data set.
Tags $\tagvar{3}$ to $\tagvar{10}$ relate to the initial position of a vehicle with respect to the ego vehicle.
The tags $\tagvar{11}$ and $\tagvar{12}$ apply if an actor is substantially slower or faster than the ego vehicle, respectively.
The remaining tags describe the longitudinal ($\tagvar{13}$ to $\tagvar{15}$) and lateral ($\tagvar{16}$ to $\tagvar{18}$) activities of vehicles surrounding the ego vehicle.
I.e., if there are \num{5} cars surrounding the ego vehicle, than the tag $\tagvar{1}$ is applied only once.

\Cref{tab:tag matrix} lists the results of the number of scenarios that contain a certain tag, i.e., the function $\tagcount{\tagsymbol}{\categorysymbol}$ that has been introduced in \cref{sec:tag coverage}.
Note that some tags are by definition part of a scenario, e.g., scenarios belonging to $\category{1}$, $\category{2}$, $\category{3}$, or $\category{6}$ always contain the tag $\tagvar{3}$.

\begin{table*}
	\centering
	\caption{$\tagcount{\tagsymbol}{\categorysymbol}$ for various tags and scenario categories, with the corresponding \acfp{sc} names listed in \cref{tab:scenario categories}.}
	\label{tab:tag matrix}
	\begin{tabular}{llrrrrrrrrrr}
    \toprule
    Symbol & Tag                                              & $\category{1}$ & $\category{2}$ & $\category{3}$ & $\category{4}$ & $\category{5}$ & $\category{6}$ & $\category{7}$ & $\category{8}$ & $\category{9}$ & $\category{10}$ \\ \otoprule
    $\tagvar{1}$ & Car                                              & \num{102111} & \num{22292} & \num{20341} & \num{5050} & \num{2992} & \num{3067} & \num{2147} & \num{819} & \num{37996} & \num{40305} \\
    $\tagvar{2}$ & Truck                                            & \num{81475} & \num{19406} & \num{17454} & \num{4234} & \num{2273} & \num{2624} & \num{1915} & \num{734} & \num{34652} & \num{31999} \\\otoprule
    $\tagvar{3}$ & Same lane in front                               & \num{102308} & \num{22296} & \num{20351} & \num{5052} & \num{1188} & \num{3069} & \num{834} & \num{480} & \num{29295} & \num{29171} \\
    $\tagvar{4}$ & Same lane rear                                   & \num{37281} & \num{11248} & \num{14377} & \num{2386} & \num{1339} & \num{1597} & \num{1006} & \num{351} & \num{23666} & \num{24913} \\
    $\tagvar{5}$ & In front left lane                               & \num{70385} & \num{17664} & \num{15934} & \num{3860} & \num{1857} & \num{2377} & \num{980} & \num{578} & \num{37850} & \num{14773} \\
    $\tagvar{6}$ & In front right lane                              & \num{49139} & \num{9190} & \num{8871} & \num{2443} & \num{2208} & \num{1187} & \num{820} & \num{476} & \num{12388} & \num{32625} \\
    $\tagvar{7}$ & At side left lane                                & \num{4952} & \num{2052} & \num{1552} & \num{228} & \num{161} & \num{205} & \num{40} & \num{17} & \num{870} & \num{1151} \\
    $\tagvar{8}$ & At side right lane                               & \num{4850} & \num{1284} & \num{1162} & \num{201} & \num{166} & \num{95} & \num{44} & \num{20} & \num{1216} & \num{1283} \\
    $\tagvar{9}$ & Rear left lane                                   & \num{32394} & \num{12730} & \num{13183} & \num{2243} & \num{1245} & \num{1750} & \num{1205} & \num{366} & \num{24979} & \num{12760} \\
    $\tagvar{10}$ & Rear right lane                                  & \num{31462} & \num{8052} & \num{8741} & \num{1777} & \num{1387} & \num{807} & \num{721} & \num{275} & \num{11063} & \num{37005} \\\otoprule
    $\tagvar{11}$ & Slower ($\Delta v < \SI{-5}{\meter\per\second}$) & \num{54750} & \num{13873} & \num{14138} & \num{4021} & \num{1348} & \num{2369} & \num{1528} & \num{591} & \num{35107} & \num{7480} \\
    $\tagvar{12}$ & Faster ($\Delta v > \SI{5}{\meter\per\second}$)  & \num{41061} & \num{9032} & \num{8046} & \num{1798} & \num{1831} & \num{957} & \num{931} & \num{403} & \num{8124} & \num{37569} \\\otoprule
    $\tagvar{13}$ & Cruising                                         & \num{102308} & \num{22296} & \num{20351} & \num{5043} & \num{2964} & \num{3051} & \num{2142} & \num{816} & \num{37935} & \num{39660} \\
    $\tagvar{14}$ & Accelerating                                     & \num{57081} & \num{22296} & \num{7652} & \num{2554} & \num{1610} & \num{1481} & \num{1260} & \num{516} & \num{21270} & \num{24039} \\
    $\tagvar{15}$ & Decelerating                                     & \num{58144} & \num{9107} & \num{20351} & \num{3419} & \num{1794} & \num{1833} & \num{1287} & \num{607} & \num{21132} & \num{24804} \\\otoprule
    $\tagvar{16}$ & Keeping lane                                     & \num{102308} & \num{22296} & \num{20351} & \num{5052} & \num{2992} & \num{3068} & \num{2156} & \num{819} & \num{38147} & \num{40307} \\
    $\tagvar{17}$ & Changing lane left                               & \num{6771} & \num{1405} & \num{1759} & \num{384} & \num{2090} & \num{2101} & \num{32} & \num{13} & \num{2545} & \num{2668} \\
    $\tagvar{18}$ & Changing lane right                              & \num{4154} & \num{1127} & \num{982} & \num{339} & \num{987} & \num{1073} & \num{12} & \num{15} & \num{1794} & \num{1741} \\
    \bottomrule
\end{tabular}
\end{table*}

\Cref{fig:results tag based} shows the tag-based coverage that results from the numbers listed in \cref{tab:tag matrix}.
It also shows how the tag-based coverage depends on the set of tags that are considered. 
For any choice of $\tagset$, we have $\tagcoverage{10}=1$, meaning that for each tag there are at least \num{10} scenarios from each \ac{sc} that contain that tag.
We still have $\tagcoverage{100}=1$ when considering tags $\tagvar{1}$, $\tagvar{2}$, and $\tagvar{10}$ to $\tagvar{14}$.
For increasing values of $\numbervar$, $\tagcoverage{\numbervar}$ starts to decrease.
To further investigate why $\tagcoverage{\numbervar}<1$, the numbers in \cref{tab:tag matrix} can be studied.
For example, given the relatively low occurrence of scenarios from \acp{sc} $\category{7}$ and $\category{8}$, the counts of the tags for these \acp{sc} are also relatively low.

\setlength{\figurewidth}{.8\linewidth}
\setlength{\figureheight}{.7\linewidth}
\begin{figure}
	\centering
\begin{tikzpicture}

\definecolor{darkblue00160}{RGB}{0,0,160}
\definecolor{darkcyan0160120}{RGB}{0,160,120}
\definecolor{darkgoldenrod2001200}{RGB}{200,120,0}
\definecolor{darkgray176}{RGB}{176,176,176}
\definecolor{darkviolet1200200}{RGB}{120,0,200}
\definecolor{lightgray204}{RGB}{204,204,204}
\definecolor{lime02200}{RGB}{0,220,0}
\definecolor{red22000}{RGB}{220,0,0}

\begin{axis}[
height=\figureheight,
legend cell align={left},
legend style={at={(.8, .98)}, anchor=north west, nodes={scale=0.7, transform shape}},
legend style={fill opacity=1, draw opacity=1, text opacity=1, draw=lightgray204},
scaled y ticks=false,
tick align=outside,
tick pos=left,
width=\figurewidth,
x grid style={darkgray176},
xlabel={$\numbervar$},
xmin=0, xmax=5,
xtick style={color=black},
xtick={0.5,1.5,2.5,3.5,4.5},
xticklabel style={align=center},
xticklabels={10,100,1000,10000,100000},
y grid style={darkgray176},
ylabel={$\tagcoverage{\numbervar}$},
ymin=0, ymax=1.05,
ytick style={color=black},
yticklabel style={/pgf/number format/fixed,/pgf/number format/precision=3}
]
\draw[draw=none,fill=darkblue00160] (axis cs:0.1,0) rectangle (axis cs:0.233333333333333,1);
\addlegendimage{ybar,ybar legend,draw=none,fill=darkblue00160}
\addlegendentry{$\tagset=\{\tagvar{1},\tagvar{2}\}$}

\draw[draw=none,fill=darkblue00160] (axis cs:1.1,0) rectangle (axis cs:1.23333333333333,1);
\draw[draw=none,fill=darkblue00160] (axis cs:2.1,0) rectangle (axis cs:2.23333333333333,0.97765);
\draw[draw=none,fill=darkblue00160] (axis cs:3.1,0) rectangle (axis cs:3.23333333333333,0.629275);
\draw[draw=none,fill=darkblue00160] (axis cs:4.1,0) rectangle (axis cs:4.23333333333333,0.2158875);
\draw[draw=none,fill=red22000] (axis cs:0.233333333333333,0) rectangle (axis cs:0.366666666666667,1);
\addlegendimage{ybar,ybar legend,draw=none,fill=red22000}
\addlegendentry{$\tagset=\{\tagvar{3},\ldots,\tagvar{9}\}$}

\draw[draw=none,fill=red22000] (axis cs:1.23333333333333,0) rectangle (axis cs:1.36666666666667,0.9645);
\draw[draw=none,fill=red22000] (axis cs:2.23333333333333,0) rectangle (axis cs:2.36666666666667,0.8341875);
\draw[draw=none,fill=red22000] (axis cs:3.23333333333333,0) rectangle (axis cs:3.36666666666667,0.45283375);
\draw[draw=none,fill=red22000] (axis cs:4.23333333333333,0) rectangle (axis cs:4.36666666666667,0.105149875);
\draw[draw=none,fill=lime02200] (axis cs:0.366666666666667,0) rectangle (axis cs:0.5,1);
\addlegendimage{ybar,ybar legend,draw=none,fill=lime02200}
\addlegendentry{$\tagset=\{\tagvar{10},\tagvar{11}\}$}

\draw[draw=none,fill=lime02200] (axis cs:1.36666666666667,0) rectangle (axis cs:1.5,1);
\draw[draw=none,fill=lime02200] (axis cs:2.36666666666667,0) rectangle (axis cs:2.5,0.9441);
\draw[draw=none,fill=lime02200] (axis cs:3.36666666666667,0) rectangle (axis cs:3.5,0.542295);
\draw[draw=none,fill=lime02200] (axis cs:4.36666666666667,0) rectangle (axis cs:4.5,0.1224785);
\draw[draw=none,fill=darkviolet1200200] (axis cs:0.5,0) rectangle (axis cs:0.633333333333333,1);
\addlegendimage{ybar,ybar legend,draw=none,fill=darkviolet1200200}
\addlegendentry{$\tagset=\{\tagvar{12},\ldots,\tagvar{14}\}$}

\draw[draw=none,fill=darkviolet1200200] (axis cs:1.5,0) rectangle (axis cs:1.63333333333333,1);
\draw[draw=none,fill=darkviolet1200200] (axis cs:2.5,0) rectangle (axis cs:2.63333333333333,0.964633333333333);
\draw[draw=none,fill=darkviolet1200200] (axis cs:3.5,0) rectangle (axis cs:3.63333333333333,0.590453333333333);
\draw[draw=none,fill=darkviolet1200200] (axis cs:4.5,0) rectangle (axis cs:4.63333333333333,0.172165);
\draw[draw=none,fill=darkgoldenrod2001200] (axis cs:0.633333333333333,0) rectangle (axis cs:0.766666666666667,1);
\addlegendimage{ybar,ybar legend,draw=none,fill=darkgoldenrod2001200}
\addlegendentry{$\tagset=\{\tagvar{15},\ldots,\tagvar{18}\}$}

\draw[draw=none,fill=darkgoldenrod2001200] (axis cs:1.63333333333333,0) rectangle (axis cs:1.76666666666667,0.890666666666667);
\draw[draw=none,fill=darkgoldenrod2001200] (axis cs:2.63333333333333,0) rectangle (axis cs:2.76666666666667,0.819433333333333);
\draw[draw=none,fill=darkgoldenrod2001200] (axis cs:3.63333333333333,0) rectangle (axis cs:3.76666666666667,0.320263333333333);
\draw[draw=none,fill=darkgoldenrod2001200] (axis cs:4.63333333333333,0) rectangle (axis cs:4.76666666666667,0.08906);
\draw[draw=none,fill=darkcyan0160120] (axis cs:0.766666666666667,0) rectangle (axis cs:0.9,1);
\addlegendimage{ybar,ybar legend,draw=none,fill=darkcyan0160120}
\addlegendentry{$\tagset=\{\tagvar{1},\ldots,\tagvar{18}\}$}

\draw[draw=none,fill=darkcyan0160120] (axis cs:1.76666666666667,0) rectangle (axis cs:1.9,0.966);
\draw[draw=none,fill=darkcyan0160120] (axis cs:2.76666666666667,0) rectangle (axis cs:2.9,0.881622222222222);
\draw[draw=none,fill=darkcyan0160120] (axis cs:3.76666666666667,0) rectangle (axis cs:3.9,0.48322);
\draw[draw=none,fill=darkcyan0160120] (axis cs:4.76666666666667,0) rectangle (axis cs:4.9,0.127867);
\end{axis}

\end{tikzpicture}
	\caption{Results of the tag-based coverage.}
	\label{fig:results tag based}
\end{figure}
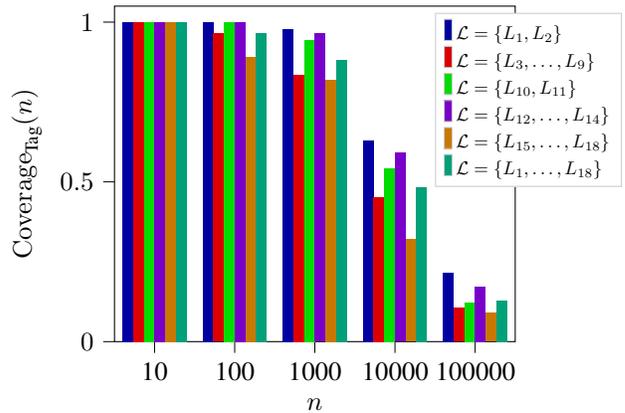

\subsection{Results time-based coverage}
\label{sec:results time based}

When calculating the time-based coverage, the time instants of each dataset containing an ego vehicle are treated separately, after which the results are combined and shown in \cref{fig:results time based}.
\Cref{fig:results time based} shows that about \SI{75}{\percent} of the time instants are covered by at least a single scenario. 
This indicates that still a substantial portion of the time instants are not covered by a scenario. 
It requires further investigation to determine if important scenario categories are missed.
In this case, in the remaining \SI{25}{\percent} of the time instants, there is no actor that complies with any of the descriptions listed in \cref{tab:scenario categories}, or simply no other actor at all.
Indeed, if we would add the scenario category ``ego vehicle has no leading vehicle'', then $\timecoverage{1}=1$.

\setlength{\figurewidth}{\linewidth}
\setlength{\figureheight}{.5\figurewidth}
\begin{figure}
	\centering
\begin{tikzpicture}

\definecolor{darkblue00160}{RGB}{0,0,160}
\definecolor{darkgray176}{RGB}{176,176,176}

\begin{axis}[
height=\figureheight,
legend style={nodes={scale=0.7, transform shape}},
scaled y ticks=false,
tick align=outside,
tick pos=left,
width=\figurewidth,
x grid style={darkgray176},
xlabel={$\numbervar$},
xmin=0.31, xmax=6.69,
xtick style={color=black},
xticklabel style={align=center},
y grid style={darkgray176},
ylabel={$\timecoverage{\numbervar}$},
ymin=0, ymax=0.8,
ytick style={color=black},
yticklabel style={/pgf/number format/fixed,/pgf/number format/precision=3}
]
\draw[draw=none,fill=darkblue00160] (axis cs:0.6,0) rectangle (axis cs:1.4,0.750014662234139);
\draw[draw=none,fill=darkblue00160] (axis cs:1.6,0) rectangle (axis cs:2.4,0.476972205277084);
\draw[draw=none,fill=darkblue00160] (axis cs:2.6,0) rectangle (axis cs:3.4,0.324293563421487);
\draw[draw=none,fill=darkblue00160] (axis cs:3.6,0) rectangle (axis cs:4.4,0.243436580921138);
\draw[draw=none,fill=darkblue00160] (axis cs:4.6,0) rectangle (axis cs:5.4,0.194757994684685);
\draw[draw=none,fill=darkblue00160] (axis cs:5.6,0) rectangle (axis cs:6.4,0.162298847309354);
\end{axis}

\end{tikzpicture}
	\caption{Results of the time-based coverage.}
	\label{fig:results time based}
\end{figure}
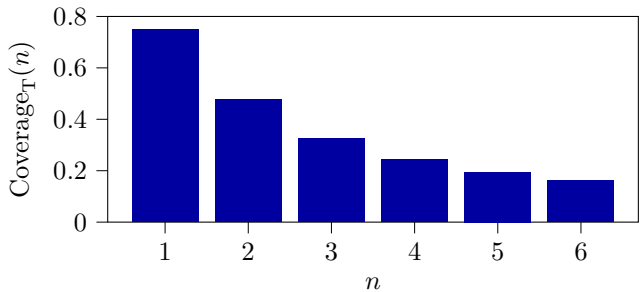

\subsection{Results actor-based coverage}
\label{sec:results actor based}

\Cref{fig:results actor based} shows the result of the actor-based coverage for different sets of actors. 
Consider an imaginary box with a certain size around the ego vehicle. 
Then, $\actorset$ contains all actors that are at some point in time within this box.
To obtain multiple values of $\actorcoverage{\actorset}$, the size of this imaginary box has been varied, as indicated in \cref{fig:results actor based}.
For $\actorscenariosset$ (the set of actors that are part of a scenario) only the main actors of a scenario, as described in \cref{tab:scenario categories}, are considered.
Note that if all actors part of a scenario would be considered, the result of the actor-based coverage would be practically similar to the time-based coverage.

When considering only actors that are ahead of the ego and in the ego vehicle's lane (blue solid line in \cref{fig:results actor based}), then $\actorcoverage{\actorset}=1$ if only vehicles that are within \SI{10}{\meter} are considered.
In all other cases, there is no full coverage. 
As with the other coverage metrics, a further investigation into the data is needed to find out why some actors are not a main actor of a scenario, even if vehicles are as near as \SI{15}{\meter}. 
In this study, these non-main actors are vehicles that are in front of the vehicle that the ego vehicle is following. 
Especially in a traffic jam, vehicles that are ahead of the leading vehicle can still be relatively close to the ego vehicle.

When changing the width of the imaginary box that contains the actors of $\actorset$, the value of $\actorcoverage{\actorset}$ drops substantially. 
This can be explained by the fact that vehicles in the adjacent lane are only considered in several occasions, namely if such a vehicle is overtaken by the ego vehicle ($\category{9}$) or if such a vehicle overtakes the ego vehicle ($\category{10})$.
Vehicles that are two lanes away from the ego vehicle are not considered as main actors for any \ac{sc}, which is why the green lines in \cref{fig:results actor based} are lower than the corresponding red lines.
It can also be noted that a substantially lower actor-based coverage is obtained if the imaginary box extends towards the back of the ego vehicle. 
This can be explained from the fact that there is only one \ac{sc} ($\category{8}$) that considers a main actor that is or could be behind the ego for the whole duration of the scenario.

\setlength{\figurewidth}{.7\linewidth}
\setlength{\figureheight}{.6\linewidth}
\begin{figure}
	\centering
\begin{tikzpicture}

\definecolor{darkblue00160}{RGB}{0,0,160}
\definecolor{darkgray176}{RGB}{176,176,176}
\definecolor{lightgray204}{RGB}{204,204,204}
\definecolor{lime02200}{RGB}{0,220,0}
\definecolor{red22000}{RGB}{220,0,0}

\begin{axis}[
height=\figureheight,
legend cell align={left},
legend style={at={(1.02, 1)}, anchor=north west, nodes={scale=0.7, transform shape}}, cells={align=left},
legend style={fill opacity=1, draw opacity=1, text opacity=1, draw=lightgray204},
scaled y ticks=false,
tick align=outside,
tick pos=left,
width=\figurewidth,
x grid style={darkgray176},
xlabel={Longitudinal distance [\si{\meter}]},
xmin=10, xmax=100,
xtick style={color=black},
xtick={10,40,70,100},
xticklabel style={align=center},
y grid style={darkgray176},
ylabel={$\actorcoverage{\actorset}$},
ymin=0.135226652102676, ymax=1.04117968323321,
ytick style={color=black},
yticklabel style={/pgf/number format/fixed,/pgf/number format/precision=3}
]
\addplot [line width=2pt, darkblue00160]
table {%
10 1
15 0.846153846153846
20 0.904761904761905
25 0.907949790794979
30 0.901129943502825
35 0.891170431211499
40 0.85625
45 0.837056504599212
50 0.809843400447427
55 0.778330019880716
60 0.741484716157205
65 0.728209191759112
70 0.706181818181818
75 0.679734219269103
80 0.661719233147805
85 0.644444444444444
90 0.634106853750675
95 0.614141414141414
100 0.59855421686747
};
\addlegendentry{Only front,\\width of \SI{3}{\meter}}
\addplot [line width=2pt, red22000]
table {%
10 0.586473429951691
15 0.547439126784215
20 0.531627576403696
25 0.518007202881152
30 0.511482254697286
35 0.502055733211512
40 0.493887530562347
45 0.481152993348115
50 0.473256601218687
55 0.457835116425425
60 0.44541231126597
65 0.435967302452316
70 0.427948717948718
75 0.415982617093192
80 0.404342857142857
85 0.395936892154744
90 0.389844559585492
95 0.379794385132463
100 0.369660148091893
};
\addlegendentry{Only front,\\width of \SI{10}{\meter}}
\addplot [line width=2pt, lime02200]
table {%
10 0.45440251572327
15 0.425733207190161
20 0.416737109044801
25 0.410506566604128
30 0.406386554621849
35 0.402976009717583
40 0.39828634604754
45 0.390381171655155
50 0.384268060836502
55 0.372387727879057
60 0.364940736119775
65 0.358208955223881
70 0.352020860495437
75 0.3441317047265
80 0.335696202531646
85 0.329059144093605
90 0.324182603331277
95 0.316809707013909
100 0.308612099644128
};
\addlegendentry{Only front,\\width of \SI{17}{\meter}}
\addplot [line width=2pt, darkblue00160, dashed]
table {%
10 0.7
15 0.593220338983051
20 0.547945205479452
25 0.492407809110629
30 0.486251808972504
35 0.467758444216991
40 0.456126482213439
45 0.437745740498034
50 0.424902289223897
55 0.404203323558162
60 0.388816644993498
65 0.378811571540266
70 0.36587982832618
75 0.354043392504931
80 0.342665855143031
85 0.333895921237693
90 0.326271186440678
95 0.316534260178749
100 0.306767271869842
};
\addlegendentry{Front and rear,\\width of \SI{3}{\meter}}
\addplot [line width=2pt, red22000, dashed]
table {%
10 0.492378048780488
15 0.431384615384615
20 0.393820500245218
25 0.366943455911427
30 0.347682119205298
35 0.331547786862137
40 0.316959921798631
45 0.301381692573402
50 0.292112950340798
55 0.27814333985417
60 0.26968311009474
65 0.260520738722374
70 0.2526790750141
75 0.243867053547877
80 0.235513090954213
85 0.229483282674772
90 0.223195476216875
95 0.21654373024236
100 0.209288299155609
};
\addlegendentry{Front and rear,\\width of \SI{10}{\meter}}
\addplot [line width=2pt, lime02200, dashed]
table {%
10 0.380208333333333
15 0.335174953959484
20 0.311817898347365
25 0.293343856879768
30 0.280534783593927
35 0.269360269360269
40 0.259324111364034
45 0.248107255520505
50 0.240356937248129
55 0.229923138086403
60 0.22338076545633
65 0.216176135067305
70 0.210064239828694
75 0.204199051175936
80 0.19784207008498
85 0.192995125473912
90 0.188079697698385
95 0.18252411312735
100 0.176406335335882
};
\addlegendentry{Front and rear,\\width of \SI{17}{\meter}}
\end{axis}

\end{tikzpicture}
	\caption{Results of the actor-based coverage.
		For the actor set $\actorset$, every actor is considered that is at some point in time within a certain longitudinal distance of the ego vehicle, varying front \SI{10}{\meter} to \SI{100}{\meter} (x-axis), and within a lateral distance, varying between \SI{1.5}{\meter} (blue), \SI{5.0}{\meter} (red), and \SI{8.5}{\meter} (green).
		For the solid lines, $\actorset$ only contain actors in front, while for the dashed lines, $\actorset$ also contain rear actors within the specified longitudinal distance.}
	\label{fig:results actor based}
\end{figure}
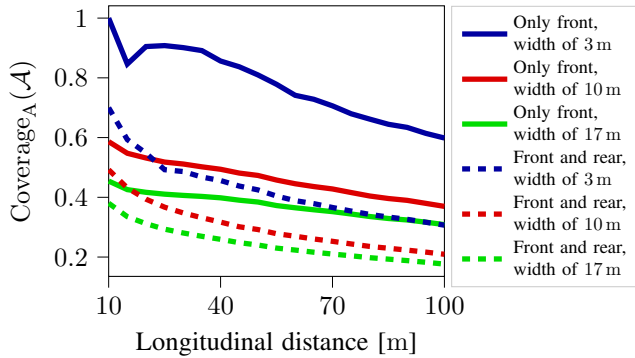

\subsection{Results actor-over-time-based coverage}
\label{sec:results time actor coverage}

\begin{figure}
	\centering
\begin{tikzpicture}

\definecolor{darkblue00160}{RGB}{0,0,160}
\definecolor{darkgray176}{RGB}{176,176,176}
\definecolor{lightgray204}{RGB}{204,204,204}
\definecolor{lime02200}{RGB}{0,220,0}
\definecolor{red22000}{RGB}{220,0,0}

\begin{axis}[
height=\figureheight,
legend cell align={left},
legend style={at={(1.02, 1)}, anchor=north west, nodes={scale=0.7, transform shape}}, cells={align=left},
legend style={fill opacity=1, draw opacity=1, text opacity=1, draw=lightgray204},
scaled y ticks=false,
tick align=outside,
tick pos=left,
width=\figurewidth,
x grid style={darkgray176},
xlabel={Longitudinal distance [\si{\meter}]},
xmin=10, xmax=100,
xtick style={color=black},
xtick={10,40,70,100},
xticklabel style={align=center},
y grid style={darkgray176},
ylabel={$\timeactorcoverage{\actorset}$},
ymin=0.104710536263114, ymax=1.02549180772825,
ytick style={color=black},
yticklabel style={/pgf/number format/fixed,/pgf/number format/precision=3}
]
\addplot [line width=2pt, darkblue00160]
table {%
10 0.983638113570741
15 0.943265579629216
20 0.948202025695058
25 0.920091698051416
30 0.904273541336918
35 0.889211329766331
40 0.857413964683284
45 0.826519940832305
50 0.793188564790914
55 0.765841865956415
60 0.729468116954396
65 0.704325356826624
70 0.689937546182422
75 0.666460450321097
80 0.64670551551398
85 0.627312162706722
90 0.616188403074736
95 0.599382308285577
100 0.583200465256087
};
\addlegendentry{Only front,\\width of \SI{3}{\meter}}
\addplot [line width=2pt, red22000]
table {%
10 0.515956292061654
15 0.386827500112804
20 0.382967123400403
25 0.399948314689704
30 0.405037800766372
35 0.418449967088153
40 0.419161837510074
45 0.41104033970276
50 0.398242545007994
55 0.388186626144539
60 0.373872663865213
65 0.364608129064373
70 0.359152098673438
75 0.350947516551463
80 0.342121680359149
85 0.333312686024313
90 0.328087313523112
95 0.320088805280864
100 0.312213996856149
};
\addlegendentry{Only front,\\width of \SI{10}{\meter}}
\addplot [line width=2pt, lime02200]
table {%
10 0.437504395156186
15 0.321370002957228
20 0.302927633883366
25 0.310542199022209
30 0.312340979164181
35 0.32246453659355
40 0.323654598496968
45 0.317902192195625
50 0.30868147890416
55 0.301169350931978
60 0.291639908975843
65 0.284966296144031
70 0.280820374126662
75 0.274864652316217
80 0.268656454175567
85 0.262356751868863
90 0.258544315394259
95 0.252540146630714
100 0.246595603866672
};
\addlegendentry{Only front,\\width of \SI{17}{\meter}}
\addplot [line width=2pt, darkblue00160, dashed]
table {%
10 0.868309260832625
15 0.726975383026338
20 0.596085637365925
25 0.535676880010664
30 0.483630230355121
35 0.471353441485595
40 0.45318379336295
45 0.432979570307349
50 0.412346261442453
55 0.398900137212745
60 0.378475778491682
65 0.365760235329527
70 0.35689904229233
75 0.347205463213747
80 0.338118200207752
85 0.32788724052461
90 0.322718551055701
95 0.315242044654944
100 0.30687777091205
};
\addlegendentry{Front and rear,\\width of \SI{3}{\meter}}
\addplot [line width=2pt, red22000, dashed]
table {%
10 0.519307622526786
15 0.368549661092132
20 0.31303053103441
25 0.290981124169063
30 0.2722913153457
35 0.268063634781873
40 0.260215593371847
45 0.248871656923593
50 0.237535285080057
55 0.229750829739322
60 0.219860885043077
65 0.213275638735998
70 0.208678963980396
75 0.203259922856787
80 0.197768797617387
85 0.192438739375627
90 0.189201343497324
95 0.184392687115999
100 0.179394812186518
};
\addlegendentry{Front and rear,\\width of \SI{10}{\meter}}
\addplot [line width=2pt, lime02200, dashed]
table {%
10 0.440879809232302
15 0.311652293894446
20 0.26093106827834
25 0.240495444219407
30 0.224414042616506
35 0.219188301278178
40 0.212042446480673
45 0.202837363137879
50 0.193763580462085
55 0.187130475631239
60 0.179666457950176
65 0.174235921639569
70 0.170311379161331
75 0.165772329795354
80 0.16141939853098
85 0.157203085995495
90 0.154450744583467
95 0.150541980564632
100 0.14656423042062
};
\addlegendentry{Front and rear,\\width of \SI{17}{\meter}}
\end{axis}

\end{tikzpicture}
	\caption{Results of the actor-over-time-based coverage.
		See \cref{fig:results actor based} for a further explanation.}
	\label{fig:results actor time based}
\end{figure}
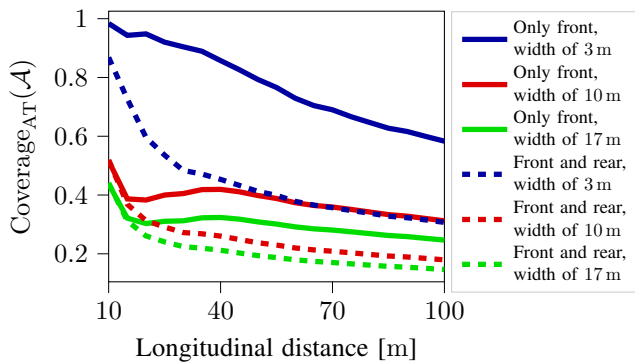

The results of the actor-over-time-based coverage are shown in \cref{fig:results actor time based}.
The lines show a similar pattern as seen in \cref{fig:results actor based}.
Especially for actors in the same lane and in front of the ego vehicle (blue lines in \cref{fig:results actor based,fig:results actor time based}), the results are nearly the same.
This indicates that those actors are generally part of a scenario as a main actor whenever they are within the imaginary box.
In the other cases (i.e., all lines except the blue solid line) the actor-over-time-based coverage is generally a bit lower, which indicates that even if the relevant actors are covered, they are not covered for the entire duration that they are within the boundaries of the imaginary box.

	\section{DISCUSSION}
\label{sec:discussion}

This work presented four different types of coverage metrics that can be used to determine whether the collected scenarios cover all relevant aspects of \iac{odd} (coverage type I) and whether the collected scenarios cover all relevant aspects that are in the data (coverage type II).
The presented coverage metrics can be helpful to decide to collect more data, start dedicated data campaigns to address parts of the \ac{odd} that are insufficiently covered, include more \acp{sc}, or decide that enough data and \acp{sc} are considered.
To decide if any of these actions is appropriate, a further investigation is needed while considering the following aspects:
\begin{itemize}
	\item As described in \cref{sec:results}, it is not always possible to achieve $\tagcoverage{\numbervar}=1$. 
	However, it may neither be necessary.
	When considering tags related to the road layout at which the scenario is taking place, it might be the case that some scenario categories do not contain hardly any scenarios with certain tags because certain type of scenarios only occur with a certain road layout.
	For example, a cut-in of another vehicle far less likely on a single-lane road.
	
	\item Even if $\timecoverage{\numbervar}=1$, it could mean that important information is not captured. 
	For example, if generic \acp{sc} are considered, such as ``driving on motorway'' and ``driving on a non-motorway road'', one can already obtain $\timecoverage{1}=1$.
	Still, it could be the case that important information is not captured by the scenarios, e.g., if the trajectory of a relevant actor is not captured.
	
	\item A high coverage type II could be a result of lots of false positives of the detected scenarios. 
	The coverage type II metrics presented in this work only check if some time instants and/or actors are covered by a scenario.
    It might be possible that a high coverage is obtained due to the false detection of scenarios.
	
	\item On the other hand, a low coverage score could be the result of many false negatives.
	I.e., if all scenarios would have been correctly detected, a higher coverage would be obtained.
\end{itemize}

Note that achieving both high coverage for type I and type II is generally difficult. 
For achieving high coverage for type II, one may need to consider many scenario categories, including scenario categories that contain only few scenarios.
As a result, it will generally be more difficult to achieve a high score for the tag-based coverage, which is a type I coverage metric.
Given that a value of \num{1} for all coverage metrics is generally difficult, it remains future work to determine --- depending on the context --- what appropriate values of the coverage are.
Note that the required coverage values may depend on the input parameters of those coverage metrics, e.g., the set of tags ($\tagset$), $\numbervar$ for $\tagcoverage{\numbervar}$ and $\timecoverage{\numbervar}$, and the conditions used to determine whether an actor is an element of $\actorset$ for $\actorcoverage{\actorset}$ and $\timeactorcoverage{\actorset}$.

\cstart
We have illustrated the presented coverage metrics by applying them on the scenarios obtained from the HighD data set. 
The experiment has shown how varying metrics can be obtained by using different tags, \acp{sc}, and values for $\numbervar$. 
While we have considered ten \acp{sc}, it is important to note that the number of \acp{sc} in a real-world application is likely to be substantially higher.
Similarly, the number of tags relevant to an \iac{odd} is expected to be higher in practice.
Next to involving more \acp{sc} and tags, future work could consider the use of additional or alternative data sets.
\cend

For increasing the trustworthiness of the data-driven, scenario-based safety assessment of \iac{ads}, not only achieving a high coverage is important, but also achieving a high completeness is important.
Here, ``completeness'' is not to be confused with ``coverage''.
Where coverage refers to the extent to which data capture aspects of interest --- in our case, the \ac{odd} --- completeness refers to the extent to which the data is free from missing values.
Similarly to coverage, two types of completeness could be considered, both aiming to answer different questions:
\begin{itemize}
	\item Type I: Do the driving data contain all relevant details of \iac{odd}?
	\item Type II: Do the collected scenarios describe all relevant details that are in the driving data?
\end{itemize}
An overview of the different types for coverage and completeness is shown in \cref{tab:coverage vs completeness}. 
An aspect considered for type I completeness could be the trajectories of actors that are simply missing from the data.
Note that in that case, a high coverage could still be obtained even though important information is missing. 
Therefore it is essential that the data is free from large omissions, i.e., high completeness type I.
Also, a high coverage could be obtained even if the collected scenarios do not contain all relevant aspects that are in the data, which would indicate insufficient completeness for type II.
In a recent work from \textcite{glasmacher2024towards}, a methodology is proposed to argue a sufficient completeness of a scenario concept, which is related to type II completeness. 
Future research is required to further address this and to develop metrics that can be used to quantify the extent of completeness.

\begin{table}
	\centering
	\caption{Overview of the different types of coverage and completeness.}
	\label{tab:coverage vs completeness}
	\begin{tabularx}{\linewidth}{lXX}
		\toprule
		& Coverage & Completeness \\\otoprule
  
		Type I & 
        Do the collected scenarios cover all relevant aspects of \iac{odd}? & 
        Do the driving data contain all relevant details of \iac{odd}? \\
        
		Type II & 
        Do the collected scenarios cover all relevant aspects that are in the driving data? & 
        Do the collected scenarios describe all relevant details that are in the driving data? \\
		\bottomrule
	\end{tabularx}
\end{table}

Both the concepts of coverage and completeness as described in \cref{tab:coverage vs completeness} refer to the \ac{odd}. 
However, for the safe deployment of \iac{ads}, the \ac{ads} must be capable of dealing safely in all operating conditions it encounters during its deployment, rather than the operating conditions it is designed for.
To describe the actual operating conditions of \iac{ads}, the term \ac{tod} is coined \autocite{ISO34503}. 
Future research should address how to measure the extent to which the \ac{odd} covers the \ac{tod}.

    \acresetall
	\section{CONCLUSIONS}
\label{sec:conclusions}

To realize the potential benefits of the deployment of \acp{ads}, the safety should be adequately ensured. 
Assessing safety can be systematically approached through scenario-based evaluations.
Data from (naturalistic) driving can serve as a source for characterizing scenarios.
This work proposes two types of coverage metrics to quantify the extent to which the data and the scenarios derived from them cover the \ac{ads}' \ac{odd}.
Coverage type I measures the extent to which the scenarios extracted from driving data cover the relevant details of \iac{odd} whereas coverage type II measures the extent to which the scenarios cover the relevant details in the driving data.
The proposed coverage metrics can be used to identify missing data or scenarios that should also be considered for the safety assessment.
Hence, the presented metrics may serve as part of the argumentation that adequate (test) scenarios have been accounted for in the safety assessment of \iac{ads}, thus aiding the type-approval process in accordance with the multi-pillar \ac{saf} outlined by the \ac{unece} \autocite{ece2021natm}.
Given that achieving \SI{100}{\%} coverage for all presented coverage metrics is not always practical or necessary, future work involves establishing suitable coverage thresholds.
Where this work outlines coverage metrics, a topic for future research is the quantification of the completeness of driving data and the identified scenarios.
Additionally, future research should be dedicated to the quantification of the degree to which the \ac{odd} encompasses the actual operating conditions of \iac{ads}.

	\addtolength{\textheight}{-10.8cm}  

	{\footnotesize\bibliographystyle{abbrvnat}
	\bibliography{bib}}

\begin{thebibliography}{29}
\providecommand{\natexlab}[1]{#1}
\providecommand{\url}[1]{\texttt{#1}}
\expandafter\ifx\csname urlstyle\endcsname\relax
  \providecommand{\doi}[1]{doi: #1}\else
  \providecommand{\doi}{doi: \begingroup \urlstyle{rm}\Url}\fi

\bibitem[Alexander et~al.(2015)Alexander, Hawkins, and
  Rae]{alexander2015situation}
R.~Alexander, H.~Hawkins, and D.~Rae.
\newblock Situation coverage --- a coverage criterion for testing autonomous
  robots.
\newblock Technical report, Department of Computer Science, University of York,
  2015.
\newblock URL \url{https://eprints.whiterose.ac.uk/88736/}.

\bibitem[Alnaser et~al.(2021)Alnaser, Sargolzaei, and
  Akba{\c{s}}]{alnaser2021autonomous}
A.~J. Alnaser, A.~Sargolzaei, and M.~I. Akba{\c{s}}.
\newblock Autonomous vehicles scenario testing framework and model of
  computation: On generation and coverage.
\newblock \emph{IEEE Access}, 9:\penalty0 60617--60628, 2021.
\newblock \doi{10.1109/ACCESS.2021.3074062}.

\bibitem[Amersbach and Winner(2019)]{amersbach2019defining}
C.~Amersbach and H.~Winner.
\newblock Defining required and feasible test coverage for scenario-based
  validation of highly automated vehicles.
\newblock In \emph{IEEE Intelligent Transportation Systems Conference (ITSC)},
  pages 425--430, 2019.
\newblock \doi{10.1109/ITSC.2019.8917534}.

\bibitem[Araujo et~al.(2023)Araujo, Mousavi, and Varshosaz]{araujo2023testing}
H.~Araujo, M.~R. Mousavi, and M.~Varshosaz.
\newblock Testing, validation, and verification of robotic and autonomous
  systems: A systematic review.
\newblock \emph{ACM Transactions on Software Engineering and Methodology},
  32\penalty0 (2):\penalty0 1--61, 2023.
\newblock \doi{10.1145/3542945}.

\bibitem[{Association for Standardization of Automation and Measuring Systems
  (ASAM)}(2021)]{openodd}
{Association for Standardization of Automation and Measuring Systems (ASAM)}.
\newblock {ASAM OpenODD}, 2021.
\newblock URL \url{https://www.asam.net/standards/detail/openodd/}.
\newblock Accessed May, 2024.

\bibitem[de~Gelder et~al.(2019)de~Gelder, Paardekooper, Op~den Camp, and
  De~Schutter]{degelder2019completeness}
E.~de~Gelder, J.-P. Paardekooper, O.~Op~den Camp, and B.~De~Schutter.
\newblock Safety assessment of automated vehicles: How to determine whether we
  have collected enough field data?
\newblock \emph{Traffic Injury Prevention}, 20\penalty0 (S1):\penalty0
  162--170, 2019.
\newblock \doi{10.1080/15389588.2019.1602727}.

\bibitem[de~Gelder et~al.(2020{\natexlab{a}})de~Gelder, Manders, Grappiolo,
  Paardekooper, Op~den Camp, and De~Schutter]{degelder2020scenariomining}
E.~de~Gelder, J.~Manders, C.~Grappiolo, J.-P. Paardekooper, O.~Op~den Camp, and
  B.~De~Schutter.
\newblock Real-world scenario mining for the assessment of automated vehicles.
\newblock In \emph{IEEE International Transportation Systems Conference
  (ITSC)}, pages 1073--1080, 2020{\natexlab{a}}.
\newblock \doi{10.1109/ITSC45102.2020.9294652}.

\bibitem[de~Gelder et~al.(2020{\natexlab{b}})de~Gelder, Op~den Camp, and
  de~Boer]{degelder2019scenariocategories}
E.~de~Gelder, O.~Op~den Camp, and N.~de~Boer.
\newblock Scenario categories for the assessment of automated vehicles.
\newblock Technical report, CETRAN, 2020{\natexlab{b}}.
\newblock URL
  \url{http://cetran.sg/wp-content/uploads/2020/01/REP200121_Scenario_Categories_v1.7.pdf}.
\newblock Version 1.7.

\bibitem[de~Gelder et~al.(2022)de~Gelder, Paardekooper, Khabbaz~Saberi,
  Elrofai, Op~den Camp, Kraines, Ploeg, and De~Schutter]{degelder2022ontology}
E.~de~Gelder, J.-P. Paardekooper, A.~Khabbaz~Saberi, H.~Elrofai, O.~Op~den
  Camp, S.~Kraines, J.~Ploeg, and B.~De~Schutter.
\newblock Towards an ontology for scenario definition for the assessment of
  automated vehicles: An object-oriented framework.
\newblock \emph{IEEE Transactions on Intelligent Vehicles}, 7\penalty0
  (2):\penalty0 300--314, 2022.
\newblock \doi{10.1109/TIV.2022.3144803}.

\bibitem[{ECE/TRANS/WP.29/2021/61}(2021)]{ece2021natm}
{ECE/TRANS/WP.29/2021/61}.
\newblock {N}ew {A}ssessment/{T}est {M}ethod for {A}utomated {D}riving ({NATM})
  -- {M}aster {D}ocument.
\newblock Technical report, World Forum for Harmonization of Vehicle
  Regulations, 2021.
\newblock URL
  \url{https://unece.org/sites/default/files/2021-04/ECE-TRANS-WP29-2021-61e.pdf}.

\bibitem[{ECE/TRANS/WP.29/2022/59/Rev.1}(2022)]{ece2022WP29}
{ECE/TRANS/WP.29/2022/59/Rev.1}.
\newblock Proposal for the 01 series of amendments to un regulation no.\ 157
  (automated lane keeping systems).
\newblock Standard, World Forum for Harmonization of Vehicle Regulations, 2022.
\newblock URL
  \url{https://unece.org/sites/default/files/2022-05/ECE-TRANS-WP.29-2022-59r1e.pdf}.

\bibitem[Elrofai et~al.(2018)Elrofai, Paardekooper, de~Gelder, Kalisvaart, and
  Op~den Camp]{elrofai2018scenario}
H.~Elrofai, J.-P. Paardekooper, E.~de~Gelder, S.~Kalisvaart, and O.~Op~den
  Camp.
\newblock Scenario-based safety validation of connected and automated driving.
\newblock Technical report, Netherlands Organization for Applied Scientific
  Research, TNO, 2018.
\newblock URL
  \url{http://publications.tno.nl/publication/34626550/AyT8Zc/TNO-2018-streetwise.pdf}.

\bibitem[{European Commission}(2022)]{ec2022typeapprovalads}
{European Commission}.
\newblock Commission implementing regulation ({EU}) 2022/1426 for the
  application of regulation ({EU}) 2019/2144 as regards uniform procedures and
  technical specifications for the type-approval of the automated driving
  system (ads) of fully automated vehicles.
\newblock \emph{Official Journal of the European Union}, L221:\penalty0 1--64,
  2022.
\newblock URL \url{https://eur-lex.europa.eu/eli/reg_impl/2022/1426/oj}.

\bibitem[Glasmacher et~al.(2023)Glasmacher, Schuldes, Weber, Wagener, and
  Eckstein]{glasmacher2023acquire}
C.~Glasmacher, M.~Schuldes, H.~Weber, N.~Wagener, and L.~Eckstein.
\newblock Acquire driving scenarios efficiently: A framework for prospective
  assessment of cost-optimal scenario acquisition.
\newblock In \emph{IEEE 26th International Conference on Intelligent
  Transportation Systems (ITSC)}, pages 1971--1976, 2023.
\newblock \doi{10.1109/ITSC57777.2023.10422027}.

\bibitem[Glasmacher et~al.(2024)Glasmacher, Weber, and
  Eckstein]{glasmacher2024towards}
C.~Glasmacher, H.~Weber, and L.~Eckstein.
\newblock Towards a completeness argumentation for scenario concepts.
\newblock In \emph{IEEE Intelligent Vehicles Symposium (IV)}, 2024.

\bibitem[Hartjen et~al.(2020)Hartjen, Philipp, Schuldt, and
  Friedrich]{hartjen2020saturation}
L.~Hartjen, R.~Philipp, F.~Schuldt, and B.~Friedrich.
\newblock Saturation effects in recorded maneuver data for the test of
  automated driving.
\newblock In \emph{13. Uni-DAS eV Workshop Fahrerassistenz und automatisiertes
  Fahren}, pages 74--83, 2020.
\newblock URL
  \url{https://www.uni-das.de/images/pdf/fas-workshop/2020/FAS_2020_HARTJEN.pdf}.

\bibitem[Hauer et~al.(2019)Hauer, Schmidt, Holzm{\"u}ller, and
  Pretschner]{hauer2019didwe}
F.~Hauer, T.~Schmidt, B.~Holzm{\"u}ller, and A.~Pretschner.
\newblock Did we test all scenarios for automated and autonomous driving
  systems?
\newblock In \emph{IEEE Intelligent Transportation Systems Conference (ITSC)},
  pages 2950--2955, 2019.
\newblock \doi{10.1109/itsc.2019.8917326}.

\bibitem[{ISO~34503}(2023)]{ISO34503}
{ISO~34503}.
\newblock Road {V}ehicles -- {T}est scenarios for automated driving systems --
  {T}axonomy for operational design domain for automated driving systems.
\newblock Standard, International Organization for Standardization, 2023.
\newblock URL \url{https://www.iso.org/standard/78952.html}.

\bibitem[{ISO~34504}(2024)]{ISO34504}
{ISO~34504}.
\newblock Road {V}ehicles -- {T}est scenarios for automated driving systems --
  {S}cenario categorization.
\newblock Standard, International Organization for Standardization, 2024.
\newblock URL \url{https://www.iso.org/standard/78953.html}.

\bibitem[Krajewski et~al.(2018)Krajewski, Bock, Kloeker, and
  Eckstein]{krajewski2018highD}
R.~Krajewski, J.~Bock, L.~Kloeker, and L.~Eckstein.
\newblock The high{D} dataset: A drone dataset of naturalistic vehicle
  trajectories on {G}erman highways for validation of highly automated driving
  systems.
\newblock In \emph{IEEE 21st International Conference on Intelligent
  Transportations Systems (ITSC)}, pages 2118--2125, 2018.
\newblock \doi{10.1109/ITSC.2018.8569552}.

\bibitem[Najm et~al.(2007)Najm, Smith, and
  Yanagisawa]{USDoT2007precrashscenarios}
W.~G. Najm, J.~D. Smith, and M.~Yanagisawa.
\newblock Pre-crash scenario typology for crash avoidance research.
\newblock Technical Report DOT HS 810 767, U.S. Department of Transportation
  Research and Innovative Technology Administration, 4 2007.
\newblock URL \url{https://rosap.ntl.bts.gov/view/dot/6281/dot_6281_DS1.pdf}.

\bibitem[{PAS 1883:2020}(2020)]{BSI1883-2020}
{PAS 1883:2020}.
\newblock Operational design domain ({ODD}) taxonomy for an automated driving
  system ({ADS}) --- specification.
\newblock Standard, The Britisch Standards Institution, 2020.

\bibitem[Piao and McDonald(2008)]{piao2008advanced}
J.~Piao and M.~McDonald.
\newblock Advanced driver assistance systems from autonomous to cooperative
  approach.
\newblock \emph{Transport Reviews}, 28\penalty0 (5):\penalty0 659--684, 2008.
\newblock \doi{10.1080/01441640801987825}.

\bibitem[Piziali(2007)]{piziali2007functional}
A.~Piziali.
\newblock \emph{Functional verification coverage measurement and analysis}.
\newblock Springer Science \& Business Media, 2007.
\newblock \doi{10.1007/b117979}.

\bibitem[Riedmaier et~al.(2020)Riedmaier, Ponn, Ludwig, Schick, and
  Diermeyer]{riedmaier2020survey}
S.~Riedmaier, T.~Ponn, D.~Ludwig, B.~Schick, and F.~Diermeyer.
\newblock Survey on scenario-based safety assessment of automated vehicles.
\newblock \emph{IEEE Access}, 8:\penalty0 87456--87477, 2020.
\newblock \doi{10.1109/ACCESS.2020.2993730}.

\bibitem[{SAE J3016}(2021)]{sae2021j3016}
{SAE J3016}.
\newblock Taxonomy and definitions for terms related to driving automation
  systems for on-road motor vehicles.
\newblock Technical report, SAE International, 4 2021.

\bibitem[Wang et~al.(2017)Wang, Liu, and Zhao]{wang2017much}
W.~Wang, C.~Liu, and D.~Zhao.
\newblock How much data are enough? {A} statistical approach with case study on
  longitudinal driving behavior.
\newblock \emph{IEEE Transactions on Intelligent Vehicles}, 2\penalty0
  (2):\penalty0 85--98, 2017.
\newblock \doi{10.1109/tiv.2017.2720459}.

\bibitem[Wasserman(2004)]{wasserman2004statistics}
L.~Wasserman.
\newblock \emph{All of Statistics: A Concise Course in Statistical Inference}.
\newblock Springer, 2004.

\bibitem[Weissensteiner et~al.(2023)Weissensteiner, Stettinger, Khastgir, and
  Watzenig]{weissensteiner2023operational}
P.~Weissensteiner, G.~Stettinger, S.~Khastgir, and D.~Watzenig.
\newblock Operational design domain-driven coverage for the safety
  argumentation of automated vehicles.
\newblock \emph{IEEE Access}, 11:\penalty0 12263--12284, 2023.
\newblock \doi{10.1109/ACCESS.2023.3242127}.

\end{thebibliography}

\end{document}